\documentclass[lettersize,journal]{IEEEtran}
\usepackage{amsmath,amsfonts}
\usepackage{algorithmic}
\usepackage{algorithm}
\usepackage{array}
\usepackage[caption=false,font=normalsize,labelfont=sf,textfont=sf]{subfig}
\usepackage{textcomp}
\usepackage{stfloats}
\usepackage{url}
\usepackage{verbatim}
\usepackage{graphicx}
\usepackage{cite}
\usepackage{bm}
\usepackage{multirow}
\usepackage{multicol}
\usepackage{booktabs}
\usepackage{caption}
\usepackage{amsfonts}
\usepackage{amsmath}
\usepackage{colortbl}
\usepackage{enumerate}
\usepackage{enumitem}

\usepackage[table]{xcolor}
\newcommand{\rfes}{\colorbox{cyan!15}{RFE-LUT-S}}
\newcommand{\rfel}{\colorbox{orange!15}{RFE-LUT-L}}

\begin{document}



\title{Receptive Field Expanded Look-Up Tables for Vision Inference: Advancing from Low-level to High-level Tasks}

\author{Xi~Zhang,~\IEEEmembership{Member,~IEEE}, and
        Xiaolin~Wu,~\IEEEmembership{Life Fellow,~IEEE}
\thanks{X.~Zhang is with the ANGEL Lab, Nanyang Technological University, Singapore. (email: xi.zhang@ntu.edu.sg).}
\thanks{X.~Wu is with the School of Computing and Artificial Intelligence, Southwest Jiaotong University, Chengdu, China (email: xwu510@gmail.com).}
}

\markboth{Journal of \LaTeX\ Class Files,~Vol.~14, No.~8, August~2021}%
{Shell \MakeLowercase{\textit{et al.}}: A Sample Article Using IEEEtran.cls for IEEE Journals}


\maketitle

\begin{abstract}
Recently, several look-up table (LUT) methods were developed to greatly expedite the inference of CNNs in a classical strategy of trading space for speed. However, these LUT methods suffer from a common drawback of limited receptive field of the convolution kernels due to the combinatorial explosion of table size. This research aims to expand the CNN receptive field with a fixed table size, thereby enhancing the performance of LUT-driven fast CNN inference while maintaining the same space complexity. To achieve this goal, various techniques are proposed. The main contribution is a novel approach of learning an optimal lattice vector quantizer that adaptively allocates the quantization resolution across data dimensions based on their significance to the inference task. In addition, the lattice vector quantizer offers an inherently more accurate approximation of CNN kernels than scalar quantizer as used in current practice. Furthermore, we introduce other receptive field expansion strategies, including irregular dilated convolutions and a U-shaped cascaded LUT structure, designed to capture multi-level contextual information without inflating table size. Together, these innovations allow our approach to effectively balance speed, accuracy, and memory efficiency, demonstrating significant improvements over existing LUT methods. 
\end{abstract}


\section{Introduction}
\label{sec:intro}
\IEEEPARstart{A}{lthough} in the past decade the CNN architecture has rapidly evolved into the arguably most popular computational model as of now, its wide deployment is hindered by the high computational complexity of CNNs, particularly for real-time applications on small, mobile end user devices.  An appealing solution to the above problem, at the CNN inference stage, is the look-up table implementation of CNNs~\cite{SRLUT,MULUT,li2024toward,SPLUT,RCLUT,DFCLUT}, i.e., adopting the classical strategy of trading space for speed. However, the limitation of the LUT realization of CNN inference is also obvious and severe. LUT limits the receptive field (RF) size of CNN kernels, because the table size grows exponentially in the receptive field size. 

To keep table sizes tractable, existing LUT methods utilize only very local neighborhoods. This design has proven effective for \emph{low-level} image restoration/enhancement  tasks~\cite{chang2004super,timofte2013anchored,timofte2015a+,lim2017enhanced,davd,agdl,mdvd,cui2025exploring,wu2025learning} 
but is inadequate for \emph{high-level} vision tasks, such as image segmentation~\cite{adams1994seeded,Shi2000,fcn,u-net,Chen2018,He2017,zhou2023nnformer,huang2025rethinking}, because the latter tasks require large RF size to aggregate long-range context and object-level cues. As of now, small-RF LUTs suffice for local detail restoration, but they hinder high-order context-based reasoning.

We reexamine the LUT design from a memory–efficiency perspective and aim to construct
a \emph{Receptive Field Expanded LUT (RFE-LUT)} without increasing the table size.
The core idea is to recast LUT design as one of optimal vector quantization (VQ); 
its objective is to approximate the input data vectors by $K$ VQ codewords, $K$ being the LUT size, while minimizing the loss of inference accuracy.  However, conventional free-form VQ methods are not suitable for fast CNN inference, as they just switch the CNN inference to another expensive problem of K nearest neighbor (KNN) search, negating the advantage of LUT~\cite{gersho2012vector}.  This is likely why all existing LUT-driven CNN inference methods adopt the simplest uniform scalar quantization (SQ) of input data to control the table size.  
To escape from the above dilemma, we find an alternative to
improve the space efficiency of LUT by replacing SQ with lattice vector quantization (LVQ) rather than the fully-fledged VQ.  The rationale of using LVQ to discretize the continuous input domain are two: 
1) the grid points of LVQ offer a more efficient Voronoi covering of vector space than 
SQ~\cite{max1960quantizing,gersho1982structure,agrell1998optimization}; 
2) the regularity of the LVQ grid makes table look up operation as simple and fast as SQ. 

Moreover, we go beyond substituting SQ with LVQ and propose a novel learning method to optimize LVQ for maximizing inference precision while being constrained by the LUT size. The method learns an optimal lattice vector quantizer whose resolution is adjusted in different data dimensions according to their importance to the inference task.  Such an optimized LVQ achieves a more efficient use of the LUT memory, which
translates to the expansion of the receptive field of the CNN kernels without increasing the table size, consequently leading to improved inference performance.
The LVQ studied in this paper for optimal design of the CNN inference LUT is similar to the classical LVQ in source coding literature in terms of efficiently tessellating a k-dimensional space by congruent k-dimensional cells. The fundamental difference between the two is in design criterion: the former is optimized for CNN inference precision whereas the latter is for minimum quantization distortion in input data space.

In addition, we also study other techniques to expand the receptive field of LUT-driven CNNs in conjunction with adaptive LVQ, such as irregular dilated convolutions and U-shaped cascaded LUT tables.
We design various irregular dilated convolution kernels that enlarge the receptive field without inflating the number of valid input pixels, thus maintaining a manageable table size. Also, we propose a U-shaped cascaded LUT structure to further extend the receptive field by fully leveraging multi-level features. This cascaded approach enables the model to capture both fine-grained details and broader contextual information, enhancing the effectiveness of LUT-based CNN inference.

Although the LUT methods can be, in principle, applied to implement any CNNs, all published works on LUT implementation of CNNs were concerned with low-level image restoration tasks.  This is apparently because of, as we pointed out ealier, the limited receptive field of LUT.  In this paper, we deliberately stress-test our RFE-LUT methods on image segmentation whose solution depends on higher order context, and demonstrate for the first time the possibility of using LUTs to boost inference speed on some high-level vision problems.  For the sake of completeness, we also report the RFE-LUT results of image super-resolution and compare them with those of the prior LUT methods. Empirically, the proposed RFE-LUT methods achieve both competitive segmentation accuracy at much higher speed and superior performance of super-resolution over the baseline LUT.  In the end, our experiments verify the key role played by enlarging the RF when extending the LUT-based inference from low-level to high-level vision tasks, under tight resource constraints on memory and computing power.

In summary, our contributions are threefold:
\begin{enumerate}
\item  We propose a lattice vector quantization (LVQ) scheme that tessellates the input space more efficiently than uniform scalar quantization and adaptively allocates per-dimension resolution under a fixed table budget, thereby improving memory efficiency and inference accuracy.

\item  We design irregular dilated convolution (IDC) kernels and a U-shaped cascaded LUT (U-LUT) architecture that expand the effective receptive field while keeping LUT dimensionality manageable, enabling integration of local details and global context without inflating storage.

\item  We demonstrate strong performance on high-level vision (nucleus and salient object segmentation) with low compute/memory cost, and show state-of-the-art gains among LUT-based methods on low-level image super-resolution, highlighting the versatility and practicality of RFE-LUT.
\end{enumerate}

\section{Background and Related Work}
\label{sec:related}

\subsection{LUT-based Image Restoration}
The Look-Up Table (LUT) is a fundamental operator in image processing~\cite{Pouli2011,Rashid2011,Mantiuk2008,Lefkimmiatis2009}, enabling rapid data retrieval via precomputed index-value mappings. Its efficiency makes it particularly useful for low-complexity operations requiring fast access.
Recently, Jo \textit{et al.} introduced SR-LUT~\cite{SRLUT}, a computationally efficient approach to super-resolution. SR-LUT trains a CNN with a limited receptive field (RF) and stores its mappings in a 4D patch-to-patch LUT, allowing direct retrieval of high-resolution (HR) patches from low-resolution (LR) inputs. However, its memory requirement grows exponentially with the RF size.

To mitigate this issue, Li \textit{et al.} proposed MuLUT~\cite{MULUT,li2024toward}, which expands the RF more efficiently by coordinating multiple LUTs with complementary indexing schemes. SPLUT~\cite{SPLUT} addresses the same challenge via cascaded LUTs, though requiring more storage overhead. Further refinements include RCLUT~\cite{RCLUT}, which employs a reconstructed convolution module to expand the RF with reduced memory, and DFC-LUT~\cite{DFCLUT}, which introduces a diagonal-first compression (DFC) scheme to optimize storage by selectively retaining high-quality information.
These advancements collectively enhance LUT-based restoration by improving RF expansion, reducing memory overhead, and maintaining computational efficiency.

More recently, several works have further advanced LUT-based restoration and broadened their applications. TinyLUT~\cite{li2024tinylut} is one such approach that aggressively reduces LUT storage footprint while preserving accuracy. It introduces a separable mapping strategy to break the LUT into smaller components, achieving over 7× reduction in storage by converting exponential growth (with kernel size) to linear. Another recent innovation is AutoLUT~\cite{xu2025autolut}, which focuses on making LUT-based networks more adaptive and learnable. Prior LUT methods used fixed sampling patterns and avoided residual connections due to value range constraints. AutoLUT addresses these limitations with two plug-and-play modules: Automatic Sampling (AutoSample) and Adaptive Residual Learning (AdaRL). The AutoSample module learns data-driven sampling patterns during training, turning static pixel selection into learnable pixel abstractions that effectively expand the receptive field without exponential memory growth. Meanwhile, AdaRL introduces modified residual connections tailored for LUT networks, allowing inter-layer information flow and feature fusion without corrupting LUT entries.

Beyond fixed-scale super-resolution, researchers have also tackled arbitrary-scale restoration with LUTs. IM-LUT~\cite{park2025lut} is a recent framework enabling continuous (non-integer) upscaling via LUTs. It trains an interpolation-mixing network (IM-Net) to blend multiple interpolation kernels (e.g. bilinear, bicubic, Lanczos, etc.), predicting content- and scale-dependent mixing weights for each local patch. LUT-based techniques have also been applied beyond traditional restoration tasks. In remote sensing, Pan-LUT~\cite{cai2025pan} adapts learned LUTs for pan-sharpening, the fusion of panchromatic and multispectral images. Another extension is in video compression: LUT-ILF~\cite{li2024loop} brings LUT learning into in-loop filtering for video codecs, which replace the conventional neural-network-based filter in VVC with a set of trained LUTs.

\subsection{Quantization in Neural Networks}
Quantization is a well-established technique for reducing the computational complexity and memory footprint of deep neural networks. By using low-precision representations (e.g., 8-bit or lower) for weights and activations instead of 32-bit floats, quantized models can achieve significant speedups and compression with minimal loss in accuracy. Early works demonstrated that neural networks could even be trained and inferred with extreme low-bit weights: BinaryConnect constrained weights to $\pm1$ during training~\cite{courbariaux2015binaryconnect}, and later Binarized Neural Networks (BNNs) extended this idea to binary activations as well, using a stochastic binarization and the straight-through estimator for backpropagation~\cite{hubara2016bnn}. XNOR-Net~\cite{rastegari2016xnor} further showed that binary-weight networks can approach full-precision accuracy on ImageNet by introducing proper rescaling factors, achieving $32\times$ memory savings and efficient bitwise operations. Pushing precision slightly higher, Ternary Weight Networks allowed weights to take values in $\{-\Delta, 0, +\Delta\}$: for example, Trained Ternary Quantization (TTQ) learned ternary weights and outperformed binary networks by retaining a zero weight option~\cite{zhu2017ttq}. These pioneering studies established that aggressive quantization of network parameters is possible, albeit with carefully designed techniques to preserve accuracy.

To minimize the accuracy gap between quantized and full-precision models, numerous learning-based quantization strategies have been developed. Many of these approaches introduce learnable quantization parameters that are optimized via gradient descent, rather than using fixed uniform quantization. For example, Choi et al. proposed PACT (Parameterized Clipping Activation), which learns the clipping threshold for activation quantization to minimize quantization error~\cite{choi2018pact}. LQ-Nets (Learning Quantization Networks) introduced a piecewise linear quantizer with learnable steps, jointly training the network and quantizer for better accuracy at low bit-widths~\cite{zhang2018lq}. Esser et al. presented Learned Step Size Quantization (LSQ), in which the step size (scale) of each quantizer is treated as a trainable variable and updated with gradients, allowing the model to automatically find optimal precision allocations for weights and activations~\cite{esser2019learned}.

\subsection{Lattice Vector Quantization (LVQ)}
Vector quantization (VQ) extends scalar quantization by mapping an entire vector
to the nearest representative in a codebook rather than quantizing each
component independently~\cite{Gray1984}.  
Lattice vector quantization (LVQ)~\cite{conway1987sphere,gibson1988lattice,ErezZamir2004}
imposes a regular geometric structure on the codebook, where representative
points form a lattice in $\mathbb{R}^d$.  
This structured design not only enables efficient indexing and decoding, but
also improves space-filling efficiency compared to uniform scalar quantization
(SQ).  

Since finding the exact nearest lattice point is NP-hard in high dimensions
\cite{micciancio2001hardness}, practical LVQ systems typically rely on
approximate decoding algorithms such as Babai’s rounding technique (BRT)
\cite{Babai1986}, which projects an input vector into the lattice coordinate
system and rounds to the nearest integer.  
Despite its simplicity, BRT often achieves near-optimal distortion performance
and has been widely adopted in coding and communication.
More recently, a line of works has explored \emph{learned lattices}:
instead of fixing the lattice basis, the basis matrix is optimized to align
the Voronoi regions with the input distribution
\cite{zhang2023lvqac,khalil2023ll,zhang2024learning,xu2025multirate,xu2025improving}.  
These learned LVQ models are especially effective in the low-bit regime where
SQ fails to preserve critical information.
Additional refinements, such as companding and linear transforms, can be incorporated to further adapt the input data to the lattice geometry~\cite{ErezZamir2004}.

In contrast to prior studies that mainly target data compression, our work
investigates a new perspective: applying LVQ to optimize \emph{inference} in
LUT-based CNNs.  
We propose a structured and task-aware quantization scheme where the quantization resolution (i.e., step size per dimension) is learned and allocated based on the relative importance of each input feature to the downstream task. This integration enables high-accuracy CNN inference through compact look-up tables, pushing the boundaries of LVQ from data representation to efficient model execution.

\section{Method}
\label{sec:method}
The proposed Receptive Field Expanded Look-Up Table (RFE-LUT) framework overcomes the limitations of conventional LUT-based CNN inference by expanding the receptive field within a fixed table size. This framework combines two main components: (1) optimized lattice vector quantization (LVQ), which improves memory efficiency by adapting quantization resolution based on the importance of input dimensions, and (2) receptive field expansion techniques, including irregular dilated convolutions and a U-shaped cascaded LUT structure, to capture both local and global context without increasing memory demands. Together, these components enable RFE-LUT to achieve high inference accuracy with efficiency, making it ideal for real-time applications on resource-limited devices. In this section, we detail each component and their synergy in enabling scalable CNN inference.

\subsection{Differentiable Lattice Vector Quantization}
\label{subsec:dlvq}

\paragraph{Preliminaries}
Let $\mathbf{B}\in\mathbb{R}^{d\times d}$ be a full-rank basis (generation) matrix, and define the lattice
\begin{equation}
    \Lambda \;=\; \bigl\{\, \mathbf{B}\mathbf{z} \;\big|\; \mathbf{z}\in\mathbb{Z}^d \,\bigr\}.
    \label{eq:lattice}
\end{equation}
Each lattice point is an integer linear combination of the columns of $\mathbf{B}$.  
Given an input vector $\mathbf{x}\in\mathbb{R}^d$, lattice vector quantization (LVQ) assigns $\mathbf{x}$ to its nearest lattice point in the Euclidean metric.  
We distinguish (i) the \emph{index} of the nearest lattice point,
\begin{equation}
    \hat{\mathbf{z}}
    \;=\;
    \arg\min_{\mathbf{z}\in\mathbb{Z}^d}
    \bigl\|\mathbf{x}-\mathbf{B}\mathbf{z}\bigr\|_2,
    \label{eq:nearest}
\end{equation}
from (ii) the \emph{nearest lattice point} (the LVQ output),
\begin{equation}
    \hat{\mathbf{x}}
    \;=\;
    Q_{\Lambda(\mathbf{B})}(\mathbf{x})
    \;=\;
    \mathbf{B}\hat{\mathbf{z}}
    \;\in\;\Lambda.
    \label{eq:quantized}
\end{equation}
Thus $\hat{\mathbf{z}}$ is the discrete lattice index, while $\hat{\mathbf{x}}$ is the quantized vector obtained by projecting $\mathbf{x}$ onto $\Lambda$ (i.e., assigning $\mathbf{x}$ to the Voronoi cell of $\hat{\mathbf{x}}$).

\paragraph{Computational challenge}
The exact problem in~\eqref{eq:nearest} is the \emph{closest vector problem} 
(CVP), which is NP-hard in the worst case~\cite{micciancio2001hardness}.  
Exact decoders such as sphere decoding become intractable when $d > 3$,
which is precisely the range relevant to LUT indexing in vision models.  
Thus, a tractable approximation is required.

\paragraph{Babai’s nearest--plane algorithm}
A widely used surrogate is Babai’s rounding algorithm~\cite{babai1986lovasz}.  
It first maps the input to lattice coordinates $\mathbf{B}^{-1}\mathbf{x}$, then 
rounds each component to the nearest integer:
\begin{equation}
    \hat{\mathbf{z}} = \left\lfloor \mathbf{B}^{-1}\mathbf{x} \right\rceil 
    \quad
    \Rightarrow
    \quad 
    \hat{\mathbf{x}} = \mathbf{B}\hat{\mathbf{z}}
    = \mathbf{B} \left\lfloor \mathbf{B}^{-1}\mathbf{x} \right\rceil,
    \label{eq:babai}
\end{equation}
where $\lfloor \cdot \rceil$ denotes element-wise rounding.  
When $\mathbf{B}$ is orthogonal or nearly orthogonal,~\eqref{eq:babai} matches 
the exact CVP solution with high probability, while retaining only $O(d)$ 
complexity.

\paragraph{Differentiability}
The main obstacle to integrating~\eqref{eq:babai} into neural networks is the 
non-differentiable rounding operator.  
We follow the uniform-noise relaxation paradigm: during back-propagation, the 
rounding is replaced by additive uniform noise 
$\mathcal{U}(-\tfrac{1}{2}, \tfrac{1}{2})$.  
This relaxation yields low-bias gradient estimates, enabling end-to-end training 
of lattice parameters jointly with the task network.

\begin{figure}[t]
    \centering
    \includegraphics[width=0.98\linewidth]{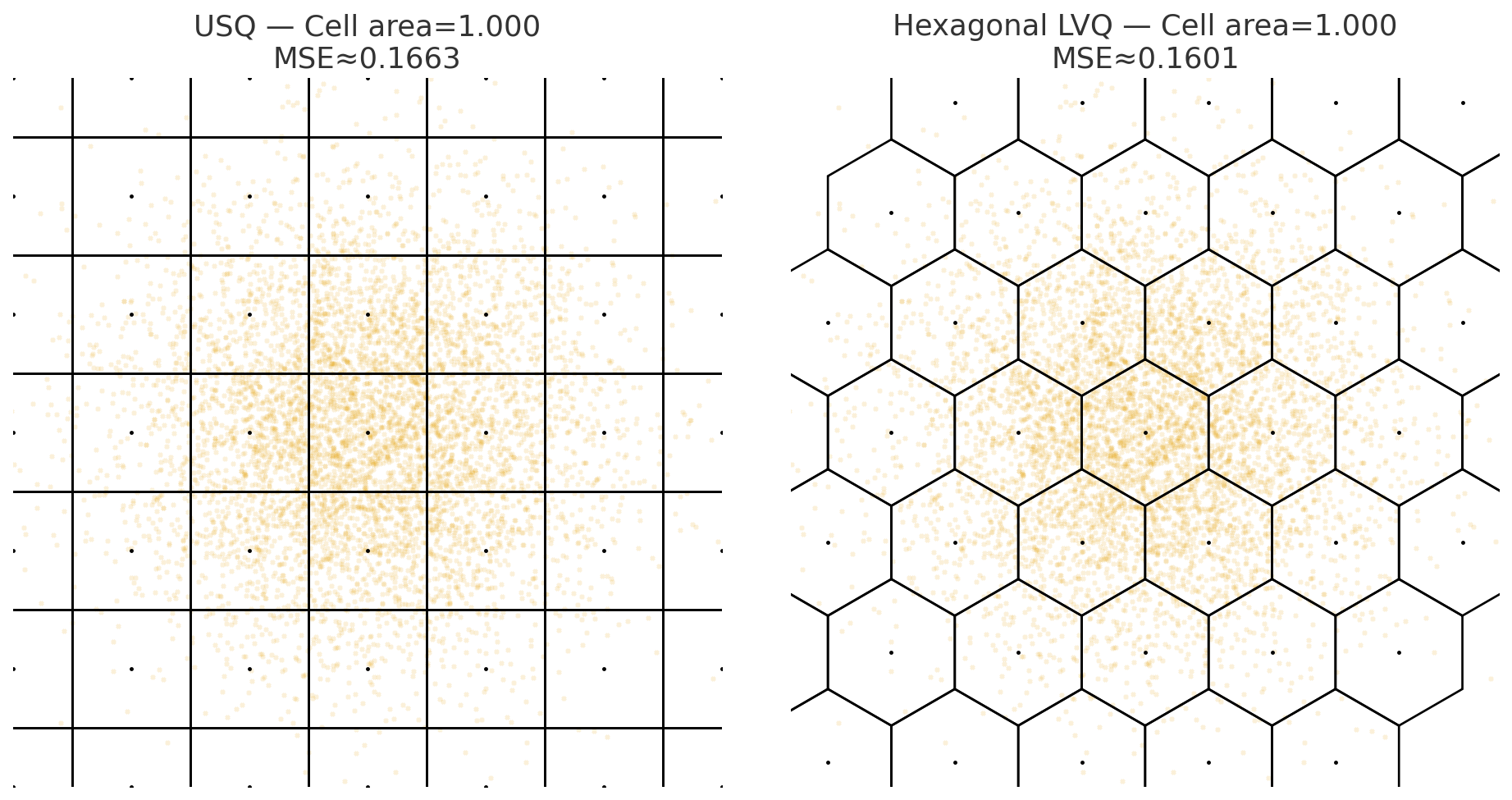}\\
    \includegraphics[width=0.98\linewidth]{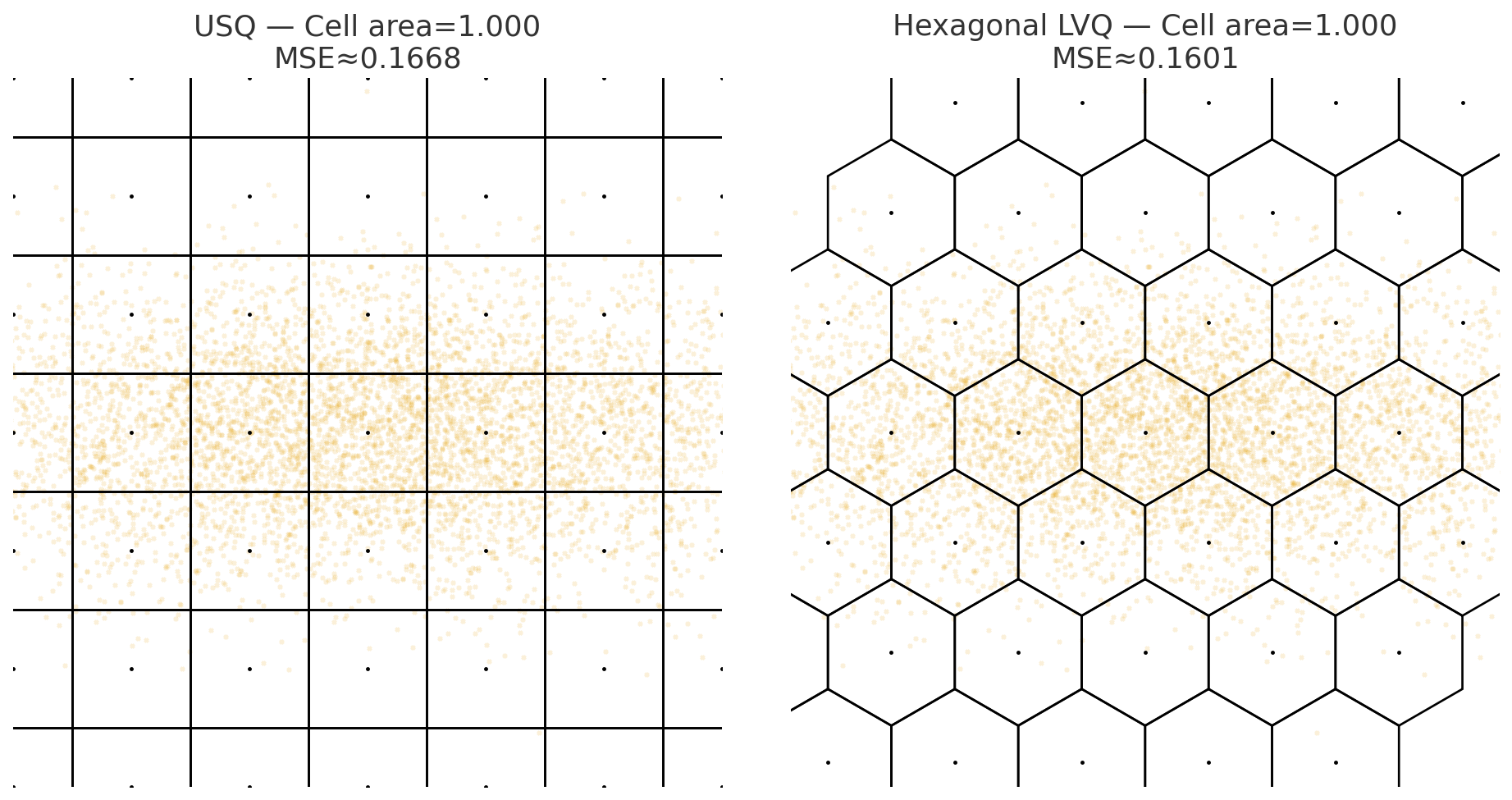}
    \caption{Comparison of uniform scalar quantization (USQ) and hexagonal lattice vector quantization (LVQ) in two dimensions. 
    The left plot shows the square lattice used in USQ, whose Voronoi cells are axis-aligned squares. 
    The right plot shows the hexagonal $A_2$ lattice used in LVQ, where the Voronoi cells form regular hexagons, achieving more efficient space filling and lower quantization error.}

    \label{fig:lvq}
\end{figure}

\paragraph{Why lattices for LUTs}
Uniform scalar quantization (SQ), the default in LUT-based CNNs such as 
SR-LUT~\cite{SRLUT}, Mu-LUT~\cite{MULUT}, and DFC-LUT~\cite{DFCLUT}, corresponds to the 
special case $\mathbf{B} = h\mathbf{I}$ with a shared step $h$.  
While efficient, SQ assigns identical resolution to all dimensions, 
ignoring their heterogeneous impact on the downstream task.  
Lattice vector quantization (LVQ), by contrast, allows 
axis-specific or oblique partitions of space, which
(i) cover the feature space more densely for a fixed codebook size 
(see Fig.~\ref{fig:lvq}), thereby reducing distortion, and 
(ii) flexibly allocate quantization resolution across dimensions.  
These properties are particularly valuable for LUT indexing, where 
compact tables must preserve fine-grained task-relevant information.

\subsection{LVQ–Driven LUT Design}
\label{subsec:lvq}

\paragraph{Hyper-rectangular lattice}
For real-time deployment we elect the \emph{hyper-rectangular} lattice,
whose basis is diagonal:
\begin{equation}
    \mathbf{B}=\operatorname{diag}(b_{1},\dots,b_{d}),\qquad b_{j}\!>\!0,
    \label{eq:diag}
\end{equation}
so that~\eqref{eq:babai} degenerates to $d$ independent scalar roundings.
The side lengths $\{b_{j}\}$, which in effect are the per-dimension step sizes, fully
control both the reconstruction fidelity and the capacity of LUTs.

\paragraph{Memory requirement}
For 8-bit activations, the number of distinct indices along dimension~$j$ is
$M_{j}=2^{8}/b_{j}+1$.  
With $m$ output values per entry (e.g.\ $m\!=\!16$ for $4\!\times\!4$
super-resolution) and $B$~bytes per value, the storage requirement is
\begin{equation}
    S(\mathbf{B}) =
    \Bigl(\prod_{j=1}^{d}M_{j}\Bigr)\,mB
    \;=\; \left(\prod_{j=1}^{d}\!\left(\frac{2^{8}}{b_j}+1\right)\right) mB .
    \label{eq:storage}
\end{equation}
Since $M_{j}$ scales inversely with $b_{j}$, finer resolution in a
dimension enlarges the table exponentially, underscoring the need for
careful, task-aware allocation of bit-budget.

\paragraph{Joint optimisation with LVQ-aware training.}
Let $\mathcal{D}=\{(\mathbf{x}_{i},y_{i})\}_{i=1}^{N}$ be the dataset and
$\theta$ the parameters of the LUT network $F_{\theta}$.
We cast the design of $\mathbf{B}$ as
\begin{equation}
\begin{aligned}
\min_{\theta,\{b_{j}\}} 
\mathcal{L}(\theta,\mathbf{B}) :=
\frac{1}{N}\!\!\sum_{i=1}^{N}\! &
\ell\!\bigl(F_{\theta}(Q_{\Lambda(\mathbf{B})}(\mathbf{x}_{i})),\,
      y_{i}\bigr)
+ \lambda\,\log S(\mathbf{B})             \\
& \text{s.t.}\quad b_{j}\ge b_{\min},\;\forall j
\end{aligned}
\label{eq:loss}
\end{equation}
where $\ell$ is the task loss (e.g.\ $\ell_{1}$ or cross-entropy), and
$\lambda$ tunes the memory–accuracy trade-off.
The logarithmic barrier smooths the otherwise discrete storage constraint,
allowing joint optimization via stochastic gradient descent.  
Empirically, we initialize $b_{j}$ from a coarse heuristic (e.g.\ equal bits
per dimension) and observe rapid convergence within a few epochs.

\paragraph{Practical impact.}
Because the basis is diagonal, the lookup latency of LVQ-LUT remains
identical to that of SQ-LUT, preserving the hallmark speed advantage.
Collectively, Differential LVQ and the above design pipeline endow
RFE-LUT with a principled mechanism to enlarge receptive fields
and shrink storage, laying the foundation for our subsequent
receptive-field–expansion modules.

\subsection{Irregular Dilated Convolution (IDC)}
\label{subsec:idc}

\begin{figure}[t]
    \centering
    \includegraphics[width=1.0\linewidth]{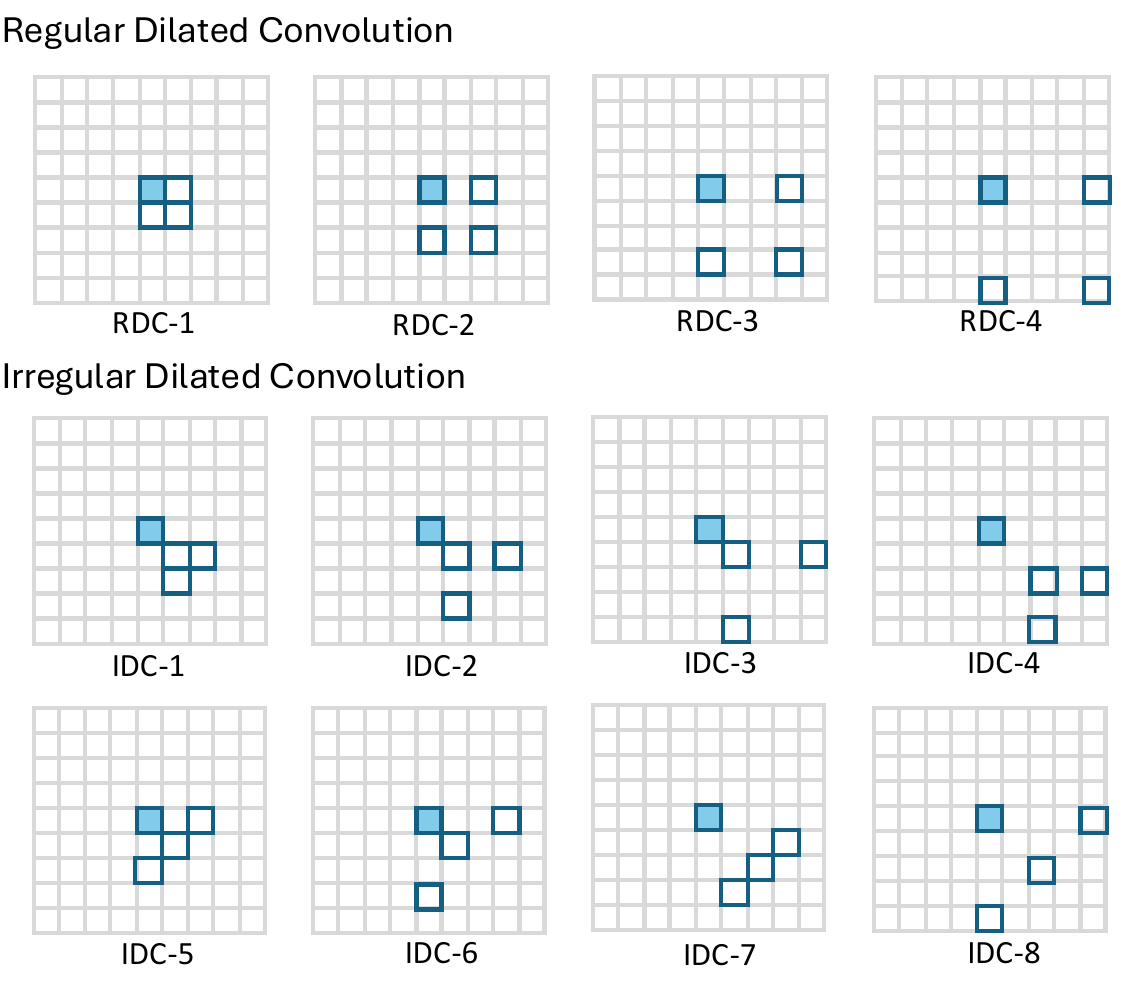}
    \caption{Comparison of regular dilated convolution (RDC) and irregular dilated convolution (IDC). In RDC, the dilation rate remains consistent across the convolution layers, resulting in uniformly spaced receptive fields. Conversely, IDC introduces variable dilation rates, enabling a flexible receptive field that captures both local and global contextual information.}
    \label{fig:idc}
\end{figure}

\paragraph{Motivation}
We start from a discrete feature map $\mathbf{F}\in\mathbb{R}^{H\times W\times C}$. 
A $k\times k$ regular dilated convolution (RDC) with rate $\delta\in\mathbb{N}_{+}$ 
samples a fixed $k^{2}$ grid of taps spaced every $\delta$ pixels along $x$ and $y$.
As $\delta$ increases, the \emph{coverage} of the receptive field expands 
quadratically (area $\Theta(\delta^{2})$), yet those taps remain locked 
to a square, axis-aligned lattice. 
This rigidity is often mismatched to long-range, \emph{directional} dependencies: 
we spend the same index budget on many locations that convey similar information,
while paying little attention to the orientations that matter.
For a LUT accelerator, where the index dimension $n$ roughly equals the number 
of active taps, our goal is therefore not to add more taps, but to place a 
\emph{small} number of them \emph{more intelligently} so that we retain the desired 
context span without inflating memory.

\paragraph{IDC definition}
To this end, we generalize RDC in two lightweight ways.
First, we allow an \emph{anisotropic} dilation vector 
$\boldsymbol{\delta}=(\delta_{x},\delta_{y})\in\mathbb{N}_{+}^{2}$, 
so that spacing along $x$ and $y$ can differ.
Second, we introduce a binary mask $M\in\{0,1\}^{k\times k}$ that activates only a 
subset of the $k^{2}$ candidate taps.
Concretely, the output at location $(p,q)$ is
\begin{equation}
\begin{aligned}
\mathbf{F}'(p,q) = &
\sum_{(m,n)\in\mathcal{R}} M_{mn}\,\mathbf{W}_{mn}\,
\mathbf{F}\!\bigl(p+\delta_{x}m,\;q+\delta_{y}n\bigr),     \\
\mathcal{R} = &
\{(m,n)\in\mathbb{Z}^{2}:\,m,n\in[-\lfloor k/2\rfloor,\lfloor k/2\rfloor]\}.
\label{eq:idc}
\end{aligned}
\end{equation}
Here $\mathbf{W}_{mn}$ denotes the per-tap weight tensor (shape determined by the 
convolution variant), and the LUT index dimension is 
$n=\sum_{(m,n)\in\mathcal{R}} M_{mn}$.
We refer to~\eqref{eq:idc} as an \emph{Irregular Dilated Convolution (IDC)}.

\paragraph{Why IDC helps}
The key is to use the same (or even smaller) index budget $n$, but distribute taps 
where they matter most for long-range interactions. 
Choosing $(\delta_{x},\delta_{y})$ with $\gcd(\delta_{x},\delta_{y})=1$ 
reduces periodic overlap and improves directional reach, 
while a sparse $M$ prunes redundant taps on the square lattice. 
As a result, IDC attains a context span comparable to a large-$\delta$ RDC, 
yet avoids the square, uniformly spaced sampling that tends to waste indices on 
near-duplicate neighborhoods. 
In short, for a given $n$, IDC offers \emph{more informative} coverage and thus 
better LUT efficiency. Please refer to Fig.~\ref{fig:idc} for the comparison of RDC and IDC.

\paragraph{Relation to deformable convolutions}
Deformable ConvNets (DCN)~\cite{dai2017deformable} learn continuous, input-dependent 
offsets and therefore require runtime interpolation and extra parameters to regress 
those offsets.
IDC, by contrast, prescribes a \emph{deterministic} lattice 
$(\boldsymbol{\delta},M)$: no additional offset parameters, 
no interpolation, and seamless compatibility with table lookup. 
IDC is thus not a special case of DCN; it is a \emph{memory-aware}, 
orientation-sensitive alternative tailored for LUT acceleration.

\begin{figure*}[t]
    \centering
    \includegraphics[width=0.98\linewidth]{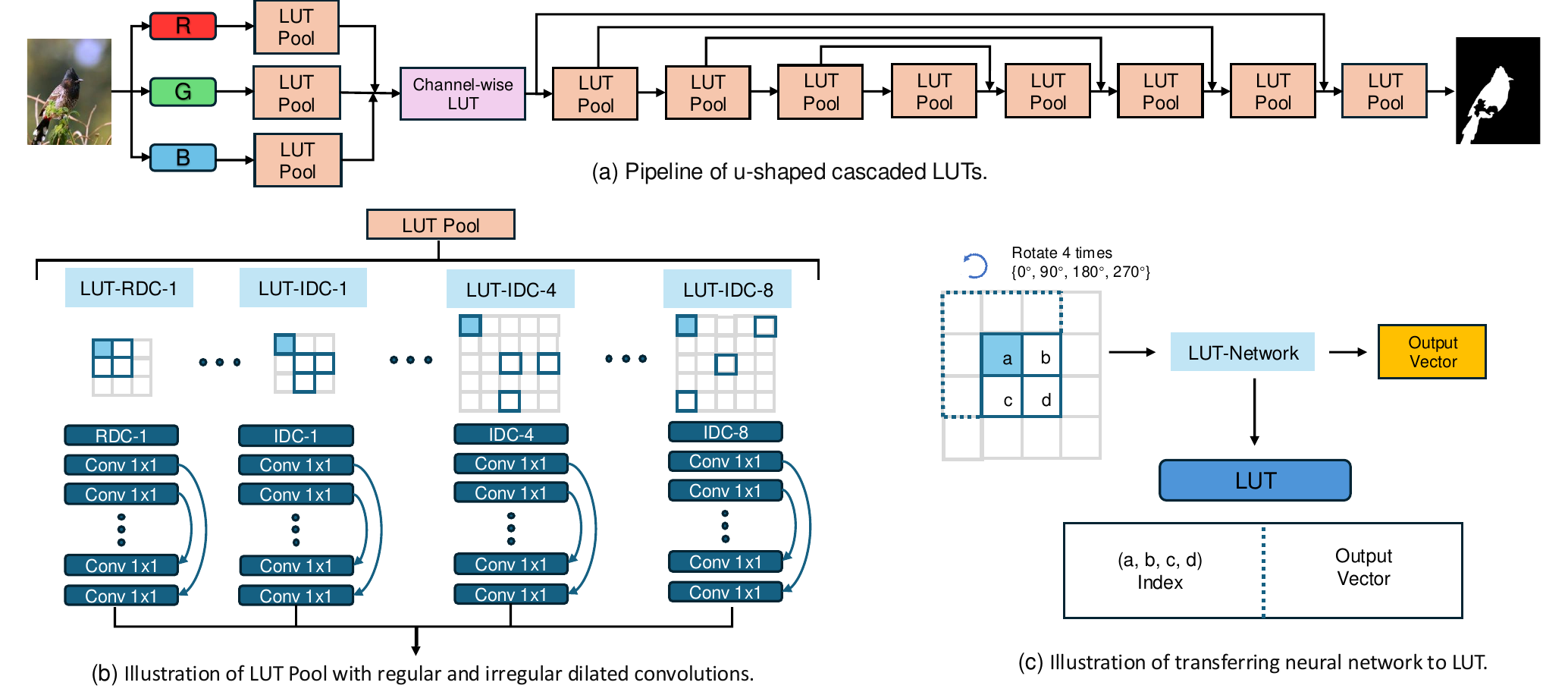}
    \caption{Overview of the proposed U-shaped cascaded LUT framework. 
    \textbf{(a)} End-to-end pipeline: RGB channels are first processed by a channel-wise LUT, then passed through a cascade of LUT pools with skip connections to aggregate long- and short-range context, producing the final prediction. 
    \textbf{(b)} Structure of a LUT pool: several parallel LUTs operate in tandem; each uses a distinct regular or irregular dilated convolution in the first layer to set the receptive field, followed by \(1\times1\) layers; their outputs are averaged. 
    \textbf{(c)} “LUT-ization” of a small CNN: responses of the trained LUT-Network are enumerated and stored in a 4-D LUT; a local \(2\times2\) window is rotated by \(\{0^\circ,90^\circ,180^\circ,270^\circ\}\) for increasing the receptive field.}
    \label{fig:u-lut}
\end{figure*}

\subsection{U-Shaped Cascaded LUTs (U\textsc{-LUT})}
\label{subsec:ulut}

\paragraph{Notation}
We organize the model into $L$ \emph{levels} of LUT \emph{pools},
denoted as $\{\mathcal{L}_{\ell}\}_{\ell=1}^{L}$.
Each pool $\mathcal{L}_{\ell}$ contains $K_{\ell}$ parallel LUTs.
The $k$-th LUT in pool $\ell$ is parameterized by $\mathcal{L}_{\ell}^{(k)}$
and is associated with a receptive\mbox{-}field sampling pattern
$\operatorname{RF}_{\ell}^{(k)}$ determined by its IDC/RDC configuration.
Let $\mathbf{F}_{\ell}\in\mathbb{R}^{H\times W\times C_{\ell}}$
be the input feature at level $\ell$ (all levels share the same spatial resolution).
The outputs of the $K_{\ell}$ branches in pool $\ell$ are combined by simple averaging,
which reduces variance among branches and makes the performance less sensitive to the branch count:
\begin{equation}
\mathbf{G}_{\ell}
=\frac{1}{K_{\ell}}\sum_{k=1}^{K_{\ell}}
\operatorname{LUT}\!\bigl(\mathbf{F}_{\ell};\,\mathcal{L}_{\ell}^{(k)}\bigr).
\label{eq:pool}
\end{equation}
This pooled output $\mathbf{G}_{\ell}$ serves as the input feature for the next level,
allowing the network to progressively enlarge the receptive field while keeping each LUT tractable.

\paragraph{Hierarchical aggregation}
We then connect pools in a U\mbox{-}shaped topology that preserves the spatial
grid at every level:
\begin{equation}
\begin{aligned}
\text{Encoder:}\quad 
& \mathbf{F}_{\ell+1}=P(\mathbf{G}_{\ell}), 
&& \ell=1,\dots,\lfloor L/2\rfloor, \\[2pt]
\text{Decoder:}\quad 
& \mathbf{F}_{\ell+1}=C\!\bigl(\mathbf{G}_{\ell},\,\mathbf{G}_{L-\ell}\bigr), 
&& \ell=\lfloor L/2\rfloor,\dots,L-1.
\label{eq:topology}
\end{aligned}
\end{equation}
Here $P$ is a $1{\times}1$ projection mapping channels \(C_\ell\!\to\!C_{\ell+1}\)
without changing resolution, and $C$ concatenates the decoder feature with its
symmetric encoder counterpart followed by a $1{\times}1$ fusion back to
\(C_{\ell+1}\).
No down/up\mbox{-}sampling is used; “U\mbox{-}shape’’ refers purely to information
flow rather than scale.

\paragraph{Memory analysis}
Let $r_{\ell,k}=|\operatorname{RF}_{\ell}^{(k)}|$ denote the number of
\emph{active LUT inputs} (i.e., effective taps) for branch $(\ell,k)$.
With IDC we enforce $r_{\ell,k}\le r_{\max}=4$, so each LUT size is bounded by
\begin{equation}
S_{\ell,k}=(2^{8-q}+1)^{\,r_{\ell,k}}\; m\,B
\;\;\le\;\;
(2^{8-q}+1)^{4}\; m\,B,
\end{equation}
where $q$ is the per\mbox{-}dimension bit\mbox{-}depth used by the table,
$2^{8-q}+1$ is the number of grid points per dimension in the interpolation
grid (for uniform quantization with boundary bins), 
$m$ is the number of output channels stored per entry (or sub\mbox{-}table
multiplicity), and $B$ is bytes per entry (e.g., FP16: 2, FP32: 4, INT8: 1).
Writing $K_{\max}=\max_{\ell} K_{\ell}$, the total footprint satisfies
\begin{equation}
S_{\text{U-LUT}}
\;\le\; L\,K_{\max}\,(2^{8-q}+1)^{4}\,m\,B,
\end{equation}
which scales linearly with depth and branch count, but crucially \emph{does not}
scale with the dilation magnitudes thanks to the bound on $r_{\ell,k}$.
In practice, skip fusions in~\eqref{eq:topology} compound diverse patterns across
levels, so the \emph{effective} receptive field grows \emph{super\mbox{-}linearly}
with $L$ (often close to multiplicative), yielding strong long\mbox{-}range context
under a strict memory cap.

\paragraph{Interpretation and takeaway}
As shown in Fig.~\ref{fig:u-lut}, encoder pools progressively re\mbox{-}encode features using \emph{complementary}
IDC/RDC sampling patterns to capture long\mbox{-}range structure; 
decoder pools align and fuse these multi\mbox{-}level cues to recover local detail,
while keeping lookup lightweight. 
In short, IDC gives each LUT an anisotropic, budgeted receptive field, and
U\textsc{-LUT} composes them hierarchically to achieve global context, 
both under
the LVQ\mbox{-}controlled memory budget of Sec.~\ref{subsec:lvq}.

\section{Experiments}
\label{sec:experiments}

In this section, we evaluate the proposed RFE-LUT framework on both high-level and low-level vision tasks to demonstrate its effectiveness in handling applications that require large receptive fields and efficient processing. For high-level tasks, we conduct experiments on medical image segmentation and salient object segmentation datasets. For low-level tasks, we assess RFE-LUT’s performance on image super-resolution, comparing it with existing LUT-based and deep CNN methods.

To balance accuracy and efficiency under different storage budgets, we instantiate two variants of our model by adjusting the regularization coefficient $\lambda$ in Eq.~(\ref{eq:loss}):
\begin{itemize}
    \item \textbf{RFE-LUT-S (Small)}: A lightweight configuration trained with a larger $\lambda$, emphasizing compactness and fast inference at minimal memory cost.
    \item \textbf{RFE-LUT-L (Large)}: A higher-capacity configuration trained with a smaller $\lambda$, prioritizing richer feature representation and prediction accuracy while maintaining efficient lookup operations.
\end{itemize}
These two variants allow us to examine the trade-off between performance and resource consumption within the same unified framework, highlighting the scalability of RFE-LUT across diverse application scenarios and hardware constraints.

\begin{table*}[t]
    \centering
    \caption{Quantitative comparison on nucleus segmentation with two benchmarks (DSB2018 and TNBC). 
    We report Hausdorff Distance (HD), Precision (PRE), Dice Similarity Coefficient (DSC), Sensitivity (SEN), and mean IoU (MIOU). 
    Classical CNN baselines (top blocks) are shown with their model sizes (MB). LUT-based methods (middle/bottom blocks) list table storage size, highlighting compactness on memory-limited devices. 
    Rows shaded as \rfes{} and \rfel{} denote our variants, which consistently reduce HD and improve other overlap-based metrics over prior LUT approaches while maintaining small storage footprints.}
    \renewcommand{\arraystretch}{1.2}
    \begin{tabular}{clcccccc}
        \hline
        \rowcolor[gray]{0.9} \textbf{Dataset} & \textbf{Method} & \textbf{Storage Size} & \textbf{HD/cm} & \textbf{PRE/(\%)} & \textbf{DSC/(\%)} & \textbf{SEN/(\%)} & \textbf{MIOU/(\%)} \\
        \hline
        \multirow{10}{*}{\textbf{DSB2018}} 
        & FCN~\cite{fcn} & 148.5MB & 2.712 & 85.41 & 80.91 & 85.15 & 81.52 \\
        & U-Net~\cite{u-net} & 32.05MB & 2.750 & 85.84 & 81.70 & 85.31 & 82.83 \\
        & Deeplabv3~\cite{deeplabv3} & 32.11MB & 2.629 & 86.68 & 82.61 & 86.60 & 84.03 \\
        & ResUNet~\cite{resunet} & 25.23MB & 2.621 & 87.35 & 82.29 & 86.95 & 83.98 \\
        & CBAM-ResUNet~\cite{zhi2022medical} & 46.23MB & 2.212 & 88.10 & 83.99 & 88.65 & 85.95 \\
        & DCA-ResUNet~\cite{zhi2022medical} & 18.67MB & 2.179 & 92.01 & 88.91 & 90.09 & 89.01 \\
        \cline{2-8}
        & SR-LUT~\cite{SRLUT} & 81.56KB & 10.821 & 55.21 & 50.09 & 54.92 & 52.47 \\
        & MuLUT~\cite{MULUT} & 489.38KB & 6.213 & 67.11 & 63.92 & 68.18 & 65.43 \\
        & DFC-LUT~\cite{DFCLUT} & 595.93KB & 5.187 & 72.38 & 68.71 & 71.66 & 70.04\\
        \cline{2-8}
        & \cellcolor{cyan!15}RFE-LUT-S 
        & \cellcolor{cyan!15}412.87KB 
        & \cellcolor{cyan!15}4.52 
        & \cellcolor{cyan!15}80.49 
        & \cellcolor{cyan!15}74.65 
        & \cellcolor{cyan!15}76.13 
        & \cellcolor{cyan!15}77.12 \\
        & \cellcolor{orange!15}RFE-LUT-L 
        & \cellcolor{orange!15}1.25MB 
        & \cellcolor{orange!15}3.46 
        & \cellcolor{orange!15}84.55 
        & \cellcolor{orange!15}79.10 
        & \cellcolor{orange!15}83.82 
        & \cellcolor{orange!15}80.34 \\
        \hline
        \multirow{11}{*}{\textbf{TNBC}} 
        & FCN & 148.5MB & 2.853 & 78.12 & 72.99 & 80.17 & 78.53 \\
        & U-Net & 32.05MB & 2.756 & 80.64 & 81.72 & 83.59 & 80.15 \\
        & Deeplabv3 & 32.11MB & 2.631 & 79.72 & 82.32 & 84.58 & 79.53 \\
        & ResUNet & 25.23MB & 2.617 & 83.41 & 83.29 & 86.93 & 83.89 \\
        & CBAM-ResUNet & 46.23MB & 2.254 & 88.53 & 83.99 & 89.21 & 86.45 \\
        & DCA-ResUNet & 18.67MB & 2.195 & 89.21 & 89.13 & 91.10 & 89.12 \\
        \cline{2-8}
        & SR-LUT~\cite{SRLUT} & 81.56KB & 11.204 & 52.63 & 48.11 & 52.08 & 50.37 \\
        & MuLUT~\cite{MULUT} & 489.38KB & 6.528 & 65.10 & 62.33 & 66.91 & 63.87 \\
        & DFC-LUT~\cite{DFCLUT} & 595.93KB & 5.392 & 69.52 & 66.24 & 68.73 & 67.15 \\
        \cline{2-8}
        & \cellcolor{cyan!20}RFE-LUT-S 
        & \cellcolor{cyan!20}412.87KB 
        & \cellcolor{cyan!20}4.75 
        & \cellcolor{cyan!20}78.01 
        & \cellcolor{cyan!20}72.39 
        & \cellcolor{cyan!20}73.72 
        & \cellcolor{cyan!20}74.63 \\
        & \cellcolor{orange!20}RFE-LUT-L 
        & \cellcolor{orange!20}1.25MB 
        & \cellcolor{orange!20}3.65 
        & \cellcolor{orange!20}83.11 
        & \cellcolor{orange!20}77.93 
        & \cellcolor{orange!20}81.32 
        & \cellcolor{orange!20}79.83 \\
        \hline
    \end{tabular}
    \label{tab:nucleus}
\end{table*}

\begin{figure*}[!h]
    \centering
    \includegraphics[width=0.98\linewidth]{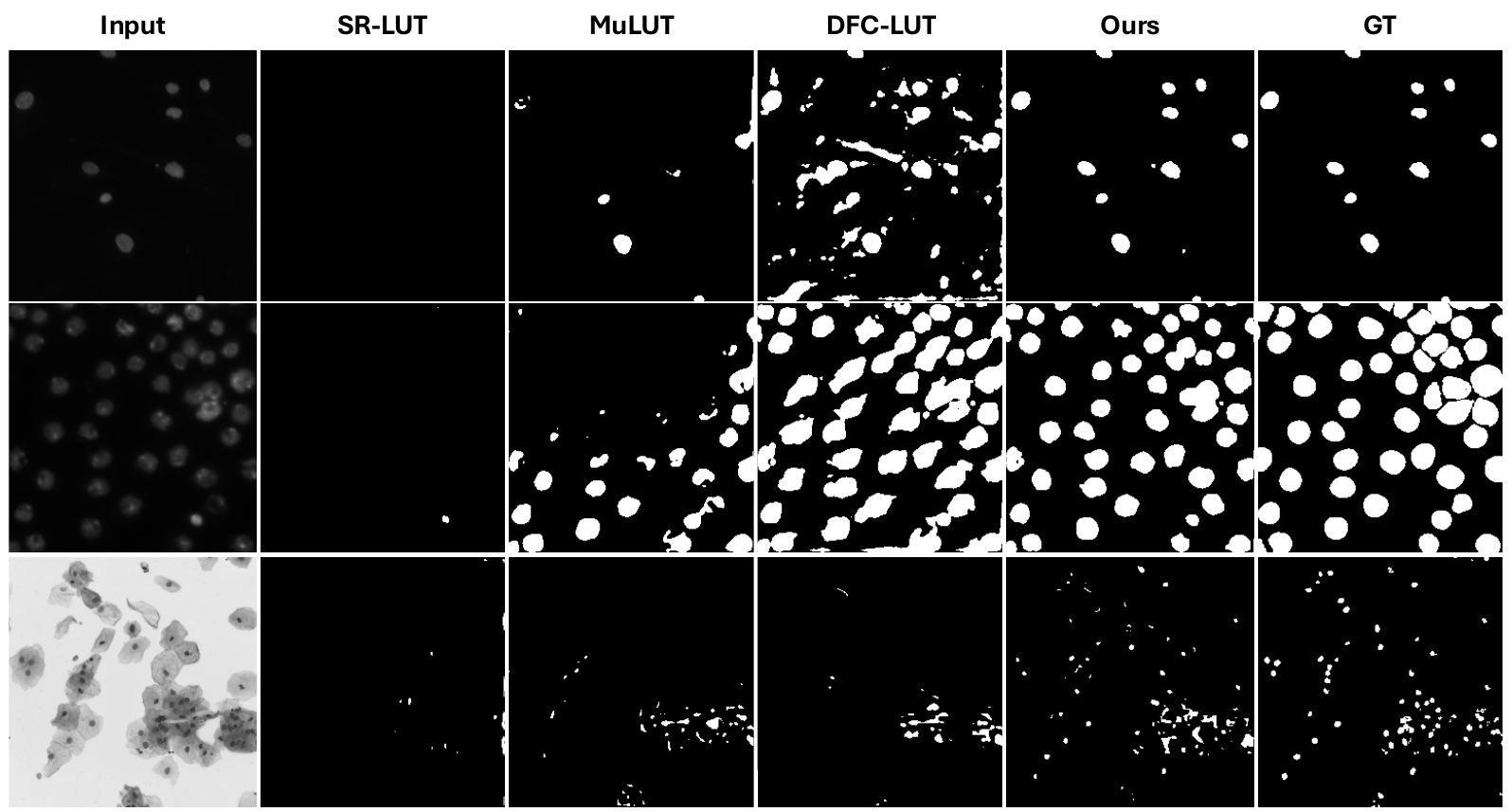}
    \caption{Qualitative comparison on DSB2018 (nucleus segmentation). 
    Each row shows the input image, predictions from prior LUT methods (SR-LUT, MuLUT, DFC-LUT), 
    our results, and the ground truth (GT). 
    Compared with baselines, RFE-LUT produces sharper nuclear boundaries, fewer false positives in background regions, and better separation of touching nuclei, especially in crowded areas and along thin structures. 
    Zoomed-in patches highlight improved boundary adherence and interior completeness, consistent with our lower HD and higher DSC/MIOU scores in the quantitative tables.}
    \label{fig:dsb2018}
\end{figure*}

\subsection{Nucleus Segmentation}
To evaluate RFE-LUT in biomedical segmentation, we conduct experiments on two challenging datasets: DSB2018~\cite{DSB2018,Kaggle2018DSB2018} and TNBC~\cite{Naylor2018TNBC}. Both datasets consist of microscopic images of cell nuclei with pixel-level annotations. DSB2018 contains images of densely packed nuclei with diverse shapes and close proximity, while TNBC includes histopathology images of triple-negative breast cancer tissues, characterized by irregular shapes, heterogeneous sizes, and high cellular density. These challenges make both datasets highly suitable for evaluating methods that require large receptive fields to capture complex contextual relationships. All images are preprocessed to $256 \times 256$, and RFE-LUT is compared against LUT-based and CNN models using Hausdorff Distance (HD), Precision (PRE), Dice Similarity Coefficient (DSC), Sensitivity (SEN), and Mean IoU (MIOU).

Table~\ref{tab:nucleus} shows that both RFE-LUT-S and RFE-LUT-L consistently outperform prior LUT-based methods on the DSB2018 and TNBC datasets across all evaluation metrics, while maintaining compact model sizes. 
On the DSB2018 dataset, the lightweight variant \textbf{RFE-LUT-S} (412.87\,KB) already achieves clear gains over the strongest baseline DFC-LUT (595.93\,KB), reducing Hausdorff Distance (HD) from $5.187$ to $4.52$ ($\downarrow12.8\%$) and improving DSC and MIOU from $68.71$/$70.04$ to $74.65$/$77.12$, respectively. This demonstrates that the proposed LVQ quantization and receptive field expansion contribute tangible performance benefits beyond mere increases in table size. The larger variant, \textbf{RFE-LUT-L} (1.25\,MB), further pushes performance boundaries, reducing HD by $33.3\%$ (from $5.187$ to $3.46$) and yielding gains of $+12.17$, $+10.39$, $+12.16$, and $+10.30$ points in PRE, DSC, SEN, and MIOU, respectively, compared to DFC-LUT. Relative to the earlier SR-LUT baseline, the improvements are even more striking, like HD drops by $68.0\%$ and DSC/MIOU increase by $+29.01/+27.87$ points, indicating much sharper boundary adherence and region consistency. While heavyweight CNNs (e.g., DCA-ResUNet) still attain the highest absolute scores, RFE-LUT-L narrows the gap considerably (e.g., MIOU $80.34$ vs.\ $89.01$) while using roughly two orders of magnitude less memory (1.25\,MB vs.\ tens of MBs), thus offering a superior accuracy–efficiency balance for deployment on resource-constrained devices.

A similar pattern is also observed on the TNBC dataset. \textbf{RFE-LUT-S} continues to deliver strong results with a sub-0.5\,MB footprint, surpassing DFC-LUT in all metrics (e.g., HD $4.75$ vs.\ $5.392$, DSC $72.39$ vs.\ $66.24$, MIOU $74.63$ vs.\ $67.15$). The full-capacity \textbf{RFE-LUT-L} further reduces HD to $3.65$ ($\downarrow33.1\%$) and increases PRE/DSC/SEN/MIOU by approximately $+12.3/+11.1/+12.2/+11.2$ points over DFC-LUT. These consistent improvements across two distinct biomedical segmentation benchmarks confirm that the proposed LVQ-driven quantization, irregular dilation, and U-shaped LUT coupling generalize robustly across domains, improving both interior-region overlap (DSC/MIOU) and contour precision (HD) in dense, heterogeneous nuclei imagery.

We present qualitative visual comparisons illustrating the segmentation results of our proposed model \textbf{RFE-LUT-L} on the DSB2018 dataset in Fig.~\ref{fig:dsb2018}. The results show that our method produces high-fidelity nucleus segmentation with precise boundary delineation and strong structural consistency, even in challenging cases involving overlapping or irregularly shaped nuclei. Compared to prior LUT-based approaches, RFE-LUT achieves noticeably sharper contours and more coherent region separation, closely matching the ground truth annotations. These visual outcomes further corroborate the quantitative improvements reported in Table~\ref{tab:nucleus}, highlighting the model’s ability to effectively integrate local detail and global context through receptive field expansion.

\begin{table*}[t]
    \centering
    \caption{Salient object detection on three benchmarks (DUTS, ECSSD, HKU-IS). 
    We report Mean Absolute Error ($M\!\downarrow$), Enhanced-alignment measure ($E_{\xi}^{m}\!\uparrow$), Structure-measure ($S_{m}\!\uparrow$), and weighted F-measure ($F_{\beta}^{w}\!\uparrow$). 
    Methods above the break are representative CNN-based models (typically $100$–$200$MB), while those below are LUT-based approaches with much smaller footprints. 
    Our \rfes{} (412.87KB) and \rfel{} (1.25MB) and  deliver consistent gains over prior LUT methods across all datasets and metrics, while narrowing the gap to heavy CNN baselines despite $100\times$–$300\times$ lower storage.}
    \renewcommand{\arraystretch}{1.0}
    \begin{tabular}{l|c|cccc|cccc|cccc}
        \hline
        \rowcolor[gray]{0.9}  
        \multirow{2}{*}{\textbf{Method}} & \textbf{Storage} & \multicolumn{4}{c}{\textbf{DUTS}} & \multicolumn{4}{c}{\textbf{ECSSD}} & \multicolumn{4}{c}{\textbf{HKU-IS}} \\
        \rowcolor[gray]{0.9} 
         & \textbf{Size} & $M\downarrow$ & $E_{\xi}^m \uparrow$ & $S_m \uparrow$ & $F_\beta^w \uparrow$ & $M\downarrow$ & $E_{\xi}^m \uparrow$ & $S_m \uparrow$ & $F_\beta^w \uparrow$ & $M\downarrow$ & $E_{\xi}^m \uparrow$ & $S_m \uparrow$ & $F_\beta^w \uparrow$ \\
        \hline
        PiCANet~\cite{picanet} &  & .04 & .915 & .863 & .812 & .035 & .953 & .916 & .908 & .031 & .951 & .905 & .89 \\
        BASNet~\cite{basnet} & & .048 & .903 & .866 & .803 & .037 & .951 & .916 & .904 & .032 & .951 & .909 & .889 \\
        CPD-R~\cite{cpd-r} & & .043 & .914 & .869 & .795 & .037 & .951 & .918 & .898 & .034 & .95 & .905 & .875 \\
        PoolNet~\cite{poolnet} & & .037 & .926 & .887 & .817 & .035 & .956 & .926 & .904 & .03 & .958 & .919 & .888 \\
        AFNet~\cite{afnet} & & .046 & .91 & .867 & .785 & .042 & .947 & .913 & .886 & .036 & .949 & .905 & .869 \\
        EGNet~\cite{egnet} & 100MB & .039 & .927 & .887 & .816 & .037 & .955 & .925 & .903 & .031 & .958 & .918 & .887 \\
        ITSD-R~\cite{itsd-r} & \multirow{2}{*}{$\mid$} & .041 & .929 & .885 & .824 & .034 & .959 & .925 & .911 & .031 & .956 & .917 & .894 \\
        MINet-R~\cite{minet-r} & & .037 & .927 & .884 & .825 & .033 & .957 & .925 & .911 & .029 & .96 & .919 & .897 \\
        LDF~\cite{ldf} & 200MB & .034 & .93 & .892 & .845 & .034 & .954 & .924 & .915 & .028 & .958 & .919 & .904 \\
        CSF-R2~\cite{csf-r2} & & .037 & .93 & .893 & .823 & .033 & .956 & .93 & .91 & .033 & .955 & .921 & .888 \\
        GateNet-R~\cite{gatenet-r} & & .04 & .928 & .885 & .809 & .034 & .952 & .925 & .894 & .033 & .955 & .915 & .88 \\
        PFSNet~\cite{pfsnet} & & .036 & .931 & .892 & .842 & .031 & .958 & .93 & .92 & .031 & .958 & .929 & .904 \\
        ICON-R~\cite{icon-r} & & .037 & .932 & .889 & .837 & .032 & .96 & .929 & .918 & .029 & .96 & .929 & .918 \\
        M3 Net-R~\cite{m3net-r} & & .036 & .937 & .897 & .849 & .029 & .962 & .931 & .919 & .026 & .966 & .929 & .913 \\
        \hline
        SR-LUT~\cite{SRLUT} & 81.56KB & .084 & .591 & .550 & .506 & .089 & .636 & .582 & .523 & .082 & .619 & .567 & .521 \\
        MuLUT~\cite{MULUT} & 489.38KB & .077 & .689 & .631 & .601 & .079 & .611 & .573 & .518 & .078 & .698 & .652 & .599 \\
        DFC-LUT~\cite{DFCLUT} & 595.93KB & .079 & .696 & .642 & .603 & .075 & .628 & .599 & .559 & .078 & .664 & .625 & .611 \\
        \hline
        \rowcolor{cyan!15}
        RFE-LUT-S & 412.87KB & .069 & .747 & .701 & .669 & .069 & .778 & .719 & .657 & .064 & .753 & .702 & .663 \\
        \rowcolor{orange!15}
        RFE-LUT-L      & 1.25MB   & .064 & .792 & .748 & .701 & .065 & .811 & .773 & .718 & .058 & .820 & .774 & .712 \\
        \hline
    \end{tabular}
    \label{tab:sod}
\end{table*}
\begin{figure*}[!h]
    \centering
    \includegraphics[width=0.98\linewidth]{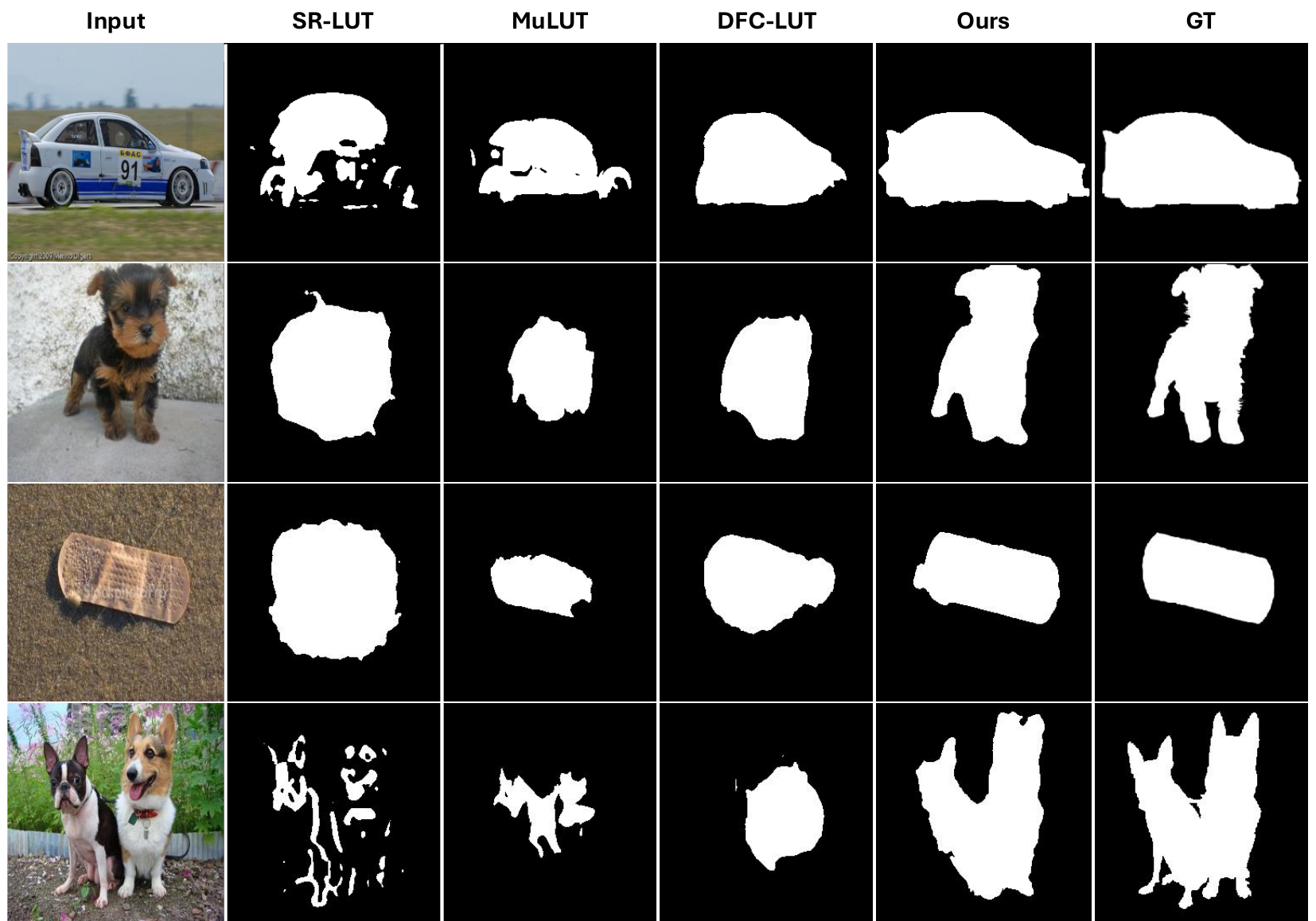}
    \caption{Visual comparison on the DUTS test set. From left to right: input image, predictions by prior LUT methods (SR-LUT, MuLUT, DFC-LUT), our results and ground-truth mask. Our results better preserve object boundaries and thin structures, suppress background clutter, and yield more complete salient regions, illustrating the benefit of receptive-field expansion with LVQ-based indexing. All methods are shown at the same resolution and without post-processing.}
    \label{fig:duts}
\end{figure*}

\subsection{Salient Object Segmentation}
To further assess the generalization of our method to high-level vision tasks, we evaluate salient object segmentation performance on three benchmark datasets: DUTS~\cite{DUTS}, ECSSD~\cite{ECSSD}, and HKU-IS~\cite{HKUIS}. 
DUTS, the largest dataset (10,553 training and 5,019 testing images), features diverse scenes with complex backgrounds, making it ideal for evaluating contextual reasoning. 
ECSSD (1,000 images) emphasizes fine structural understanding, while HKU-IS (4,447 images) contains multiple salient objects in cluttered scenes, demanding simultaneous global and local context awareness.
All images are resized to $256 \times 256$, and models are evaluated using four standard metrics: mean absolute error (MAE), E-measure ($E_{\xi}^{m}$), S-measure ($S_m$), and weighted F-measure ($F_{\beta}^{w}$).

As shown in Table~\ref{tab:sod}, \textbf{RFE-LUT-S} and \textbf{RFE-LUT-L} both achieve substantial improvements over previous LUT-based methods across all datasets and evaluation metrics. 
The compact \textbf{RFE-LUT-S} (412.87\,KB) already surpasses DFC-LUT (595.93\,KB) with consistent gains. For example, on DUTS, $E_{\xi}^{m}$ rises from $0.696$ to $0.747$, and $S_m$ from $0.642$ to $0.701$, which demonstrates that receptive field expansion and LVQ quantization yield tangible benefits even under strict memory constraints. 
The higher-capacity \textbf{RFE-LUT-L} (1.25\,MB) further elevates accuracy, achieving $E_{\xi}^{m}$/$S_m$/$F_{\beta}^{w}$ of $0.792/0.748/0.701$ on DUTS, surpassing DFC-LUT by large margins while maintaining a compact model size. 
Similar trends are observed on ECSSD and HKU-IS, where RFE-LUT-L attains consistently higher alignment-based metrics ($E_{\xi}^{m}$, $S_m$) and overlap-based scores ($F_{\beta}^{w}$), indicating improved consistency between predicted and ground-truth saliency maps. 
Although CNN-based models still report slightly higher absolute performance, RFE-LUT-L narrows the gap significantly while achieving over $100\times$ reduction in storage, demonstrating an excellent trade-off between performance and efficiency.

We further present qualitative comparisons in Fig.~\ref{fig:duts}, showcasing visual results of \textbf{RFE-LUT-L} on the DUTS dataset. 
RFE-LUT-L effectively identifies and segments salient objects with sharper boundaries and more coherent region coverage, even under challenging conditions such as low contrast, background clutter, or overlapping objects. 
Unlike previous LUT-based approaches that often yield fragmented or incomplete object masks, our method preserves fine structures and produces visually consistent results closely aligned with the ground truth. 
These visual and quantitative results jointly demonstrate the effectiveness of receptive field expansion and LVQ-aware quantization in enabling LUT-based inference to handle complex high-level vision tasks.

\subsection{Image Super-Resolution}

\begin{table*}[t]
    \centering
    \caption{Quantitative comparison of PSNR/SSIM (higher is better) and storage size for $\times4$ super-resolution on five benchmarks (Set5, Set14, BSDS100, Urban100, Manga109). We compare classical methods, CNN baselines, and LUT-based models. Rows shaded as \rfes{} and \rfel{} denote our variants, which deliver consistent gains over prior LUT approaches while remaining compact, and approach CNN performance with an order-of-magnitude smaller storage footprint.}
    \renewcommand{\arraystretch}{1.2}
    \begin{tabular}{clcccccc}
        \hline
        \rowcolor[gray]{0.9} & \textbf{Method} & \textbf{Storage Size} & \textbf{Set5} & \textbf{Set14} & \textbf{BSDS100} & \textbf{Urban100} & \textbf{Manga109} \\
        \hline
        \multirow{4}{*}{\textbf{Classical}} & Bicubic & - & 28.42/0.8101 & 26.00/0.7023 & 25.96/0.6672 & 23.14/0.6574 & 24.91/0.7871 \\
        & NE + LLE~\cite{chang2004super} & 1.434MB & 29.62/0.8404 & 26.82/0.7346 & 26.49/0.6970 & 23.84/0.6942 & 26.10/0.8195 \\
        & ANR~\cite{timofte2013anchored} & 1.434MB & 29.70/0.8422 & 26.86/0.7368 & 26.52/0.6992 & 23.89/0.6964 & 26.18/0.8214 \\
        & A+~\cite{timofte2015a+} & 15.17MB & 30.27/0.8602 & 27.30/0.7498 & 26.73/0.7088 & 24.33/0.7189 & 26.91/0.8480 \\
        \hline
        \multirow{2}{*}{\textbf{DNN}} & RRDB~\cite{wang2018esrgan} & 63.942MB & 32.68/0.8999 & 28.88/0.7891 & 27.82/0.7444 & 27.02/0.8146 & 31.57/0.9185 \\
        & EDSR~\cite{lim2017enhanced} & 164.396MB & 32.46/0.8968 & 28.80/0.7876 & 27.71/0.7420 & 26.64/0.8033 & 31.02/0.9148 \\
        \hline
        \multirow{5}{*}{\textbf{LUT}} & SR-LUT~\cite{SRLUT} & 1.274MB & 29.94/0.8524 & 27.18/0.7416 & 26.59/0.6999 & 24.09/0.7053 & 26.94/0.8454 \\
        & MuLUT~\cite{MULUT} & 4.062MB & 30.60/0.8653 & 27.60/0.7541 & 26.86/0.7110 & 24.46/0.7194 & 27.90/0.8633 \\
        & DFC-LUT~\cite{DFCLUT} & 2.018MB & 31.05/0.8755 & 27.88/0.7632 & 27.08/0.7190 & 24.81/0.7357 & 28.58/0.8779 \\
        \cline{2-8}
        & \cellcolor{cyan!15}RFE-LUT-S 
        & \cellcolor{cyan!15}1.625MB 
        & \cellcolor{cyan!15}31.22/0.8788 
        & \cellcolor{cyan!15}28.09/0.7702 
        & \cellcolor{cyan!15}27.19/0.7204 
        & \cellcolor{cyan!15}25.20/0.7511 
        & \cellcolor{cyan!15}28.74/0.8858 \\
        & \cellcolor{orange!15}RFE-LUT-L 
        & \cellcolor{orange!15}6.392MB 
        & \cellcolor{orange!15}31.84/0.8891 
        & \cellcolor{orange!15}28.59/0.7785 
        & \cellcolor{orange!15}27.41/0.7318 
        & \cellcolor{orange!15}25.92/0.7896 
        & \cellcolor{orange!15}29.41/0.8931 \\
        \hline
    \end{tabular}
    \label{tab:sr}
\end{table*}

For low-level vision tasks, we evaluate our method on image super-resolution following the standard protocol of prior LUT-based studies~\cite{SRLUT,MULUT,DFCLUT}. The same benchmark datasets (Set5, Set14, BSD100, Urban100, and Manga109), evaluation metrics (PSNR and SSIM), and $\times4$ upscaling factor are adopted to ensure a fair comparison.

As summarized in Table~\ref{tab:sr}, both \textbf{RFE-LUT-S} and \textbf{RFE-LUT-L} consistently outperform previous LUT-based methods across all datasets, achieving superior PSNR/SSIM scores with compact model sizes. The lightweight variant, \textbf{RFE-LUT-S} (1.625\,MB), already exceeds DFC-LUT (2.018\,MB) on all benchmarks, yielding notable improvements on complex datasets such as Urban100 and Manga109, where fine-grained texture reconstruction is most challenging. The higher-capacity \textbf{RFE-LUT-L} (6.392\,MB) further improves performance, reaching 31.84\,dB/0.8891 on Set5 and 25.92\,dB/0.7896 on Urban100, representing new state-of-the-art results among LUT-based approaches.

These results validate the effectiveness of our lattice vector quantization and receptive field expansion strategies in modeling complex image structures. Despite its small storage footprint compared to deep CNNs, RFE-LUT-L captures richer contextual dependencies and delivers superior perceptual quality, while RFE-LUT-S provides an even more compact alternative for resource-constrained deployments. Together, they demonstrate that RFE-LUT generalizes effectively from high-level segmentation tasks to low-level restoration, unifying efficiency and quality under a single LUT-based framework.

\subsection{Ablation Studies}
\begin{table*}[t]
    \centering
    \caption{Ablation on DSB2018 quantifying the contribution of each component in RFE-LUT. We progressively add Lattice Vector Quantization (LVQ), Irregular Dilated Convolution (IDC), and the U-shaped cascaded LUT (U-LUT) to a baseline without any of them. Metrics include Hausdorff Distance (HD, $\downarrow$), Precision (PRE, $\uparrow$), Dice (DSC, $\uparrow$), Sensitivity (SEN, $\uparrow$), and mIoU ($\uparrow$). The full model (LVQ+IDC+U-LUT) achieves the best performance across all metrics.}
    \renewcommand{\arraystretch}{1.2}
    \label{tab:ablation_dsb2018}
    \begin{tabular}{lcccccc}
        \hline
        \rowcolor[gray]{0.9} \textbf{Method} & \textbf{HD} $\downarrow$ & \textbf{PRE (\%)} $\uparrow$ & \textbf{DSC (\%)} $\uparrow$ &
        \textbf{SEN (\%)} $\uparrow$ & \textbf{MIOU (\%)} $\uparrow$ \\
        \hline
        Baseline (w/o LVQ, w/o IDC, w/o U-LUT) & 5.87 & 72.35 & 65.12 & 68.41 & 67.08 \\
        + LVQ Only & 4.92 & 76.84 & 70.56 & 73.29 & 71.63 \\
        + LVQ + IDC & 3.89 & 80.47 & 75.23 & 78.12 & 74.85 \\
        + LVQ + U-LUT & 4.11 & 78.34 & 73.14 & 75.62 & 73.04 \\
        \rowcolor{orange!15} \textbf{+ LVQ + IDC + U-LUT (Full Model)} & \textbf{3.46} & \textbf{84.55} & \textbf{79.10} & \textbf{83.82} & \textbf{80.34} \\
        \hline
    \end{tabular}
    \label{tab:ablation}
\end{table*}

To quantify the contribution of each component of the proposed RFE-LUT, we conduct controlled ablations on DSB2018 under a fixed LUT budget and identical training protocol (optimizer, schedule, data aug.). Rather than “removing everything but one,” we adopt a cumulative design that mirrors how the full model is built and used in practice:

\begin{itemize}
\item \textbf{Baseline}~(w/o LVQ, w/o IDC, w/o U\mbox{-}LUT): a minimal LUT backbone using uniform scalar quantization (SQ), regular $3{\times}3$ receptive field realization, and no cross-level coupling.
\item \textbf{+\,LVQ only}: replace SQ with our hyper\mbox{-}rectangular lattice vector quantization and LVQ-aware training, keeping the same receptive field and architecture otherwise. This isolates the effect of task-aware quantization.
\item \textbf{+\,LVQ\,+\,IDC}: on top of LVQ, enable the irregular dilated convolution patterns in the first layer of each LUT (subsequent layers remain $1{\times}1$). This tests targeted receptive-field expansion at constant table size.
\item \textbf{+\,LVQ\,+\,U\mbox{-}LUT}: on top of LVQ (without IDC), add the U-shaped cascaded LUT coupling with skip connections across levels to assess the benefit of multi-level feature aggregation without changing resolution.
\item \textbf{Full (LVQ\,+\,IDC\,+\,U\mbox{-}LUT)}: combine all components.
\end{itemize}

All variants keep the same index dimensionality per LUT and the same storage per entry, ensuring differences stem from the \emph{component under test}. We report Hausdorff Distance (HD$\downarrow$), Precision (PRE$\uparrow$), Dice (DSC$\uparrow$), Sensitivity (SEN$\uparrow$), and mean IoU (MIOU$\uparrow$). 

The ablation studies presented in Table~\ref{tab:ablation} demonstrates the individual and combined contributions of the proposed components, Lattice Vector Quantization (LVQ), Irregular Dilated Convolutions (IDC), and the U-shaped LUT structure (U-LUT), to the performance of the RFE-LUT framework on the DSB2018 dataset. Starting from the baseline model with none of these enhancements, we observe a substantial improvement when LVQ is introduced, with a decrease in Hausdorff Distance (HD) from 5.87 to 4.92 and increases across all other metrics, indicating that LVQ alone provides more efficient memory utilization and precise segmentation. Adding IDC alongside LVQ further improves segmentation accuracy, as shown by a significant reduction in HD to 3.89 and enhanced Dice Similarity Coefficient (DSC) and Mean Intersection over Union (MIOU), suggesting that IDC effectively expands the receptive field to capture broader spatial context. The inclusion of the U-LUT structure, either with LVQ alone or with both LVQ and IDC, yields similar benefits, showing that the multi-level feature integration in U-LUT contributes to improved sensitivity (SEN) and precision (PRE). The full model, combining LVQ, IDC, and U-LUT, achieves the best results across all metrics, particularly with a low HD of 3.46 and high DSC of 79.10\%, underscoring that these components work synergistically to maximize segmentation accuracy and boundary precision, making the RFE-LUT framework highly effective for complex biomedical segmentation tasks.

\subsection{Running Time Analysis}
\label{subsec:runtime}
Table~\ref{tab:runtime} reports the inference time required to generate a $1280 \times 720$ HD image with $\times4$ super-resolution across different categories of methods. 
Interpolation-based approaches (e.g., bicubic) are the fastest but deliver poor reconstruction quality due to their lack of learned priors. 
Classical sparse-coding methods such as A+~\cite{timofte2015a+} and ANR~\cite{timofte2013anchored} achieve moderate visual fidelity but incur high computational cost, making them impractical for real-time scenarios.

In contrast, LUT-based methods achieve a superior balance between inference speed and visual quality. 
Our compact variant, \textbf{RFE-LUT-S}, completes the HD upscaling in \textbf{220\,ms}, outperforming previous LUT models such as SR-LUT (137\,ms), MuLUT (228\,ms), and DFC-LUT (318\,ms) in accuracy while maintaining real-time feasibility. 
Besides, our larger variant, \textbf{RFE-LUT-L}, runs at \textbf{454\,ms}, still an order of magnitude faster than deep CNN models like RRDB~\cite{wang2018esrgan} (3,104\,ms) and EDSR~\cite{lim2017enhanced} (4,880\,ms). 

These results highlight that RFE-LUT’s receptive-field expansion and LVQ optimization incur negligible runtime overhead relative to baseline LUT methods while substantially improving reconstruction quality. 
This efficiency makes RFE-LUT particularly attractive for deployment on mobile or embedded devices, where low latency and limited compute resources are critical.

\begin{table}[t]
    \centering
    \caption{End-to-end runtime (milliseconds) to produce a $1280\times720$ image via $\times4$ super-resolution on mobile/PC. 
    LUT-based methods deliver sizable speedups over heavy-duty DNNs, and our variants \rfes{} and \rfel{} achieve real-time performance on mobile while retaining high restoration quality.
    }
    \renewcommand{\arraystretch}{1.2}
    \begin{tabular}{clcr}
        \hline
        \rowcolor[gray]{0.9} & \textbf{Method} & \textbf{Platform} & \textbf{RunTime (ms)} \\
        \hline
        \multirow{3}{*}{\textbf{Interpolation}}
        & Nearest  & Mobile & 9 \\
        & Bilinear & Mobile & 20 \\
        & Bicubic  & Mobile & 97 \\
        \hline
        \multirow{3}{*}{\textbf{Classical}} 
        & NE + LLE & PC & 4,687 \\
        & ANR      & PC & 1,260 \\
        & A+        & PC & 1,151 \\
        \hline
        \multirow{2}{*}{\textbf{DNN}}
        & RRDB    & Mobile & 3,104 \\
        & EDSR  & Mobile &  4,880 \\
        \hline
        \multirow{5}{*}{\textbf{LUT}} 
        & SR-LUT~\cite{SRLUT}         & Mobile & 137 \\
        & MuLUT~\cite{MULUT}         & Mobile & 228 \\
        & DFC-LUT~\cite{DFCLUT}    & Mobile & 318 \\
        \cline{2-4}
        & \cellcolor{cyan!20}RFE-LUT-S & \cellcolor{cyan!20}Mobile & \cellcolor{cyan!20}220 \\
        & \cellcolor{orange!20}RFE-LUT-L & \cellcolor{orange!20}Mobile & \cellcolor{orange!20}454 \\
        \hline
    \end{tabular}
    \label{tab:runtime}
\end{table}

\subsection{Limitations}
Despite its favorable accuracy–efficiency trade-off, RFE-LUT has several limitations.
First, the learned lattice (diagonal hyper-rectangular) and its quantization resolutions are task-specific: when the data distribution or objective changes, LVQ parameters and LUT contents typically require re-optimization, which adds engineering overhead compared to end-to-end CNNs that adapt weights directly. Second, IDC uses fixed irregular dilation patterns to preserve table-lookup compatibility, which means it cannot adapt spatial sampling at runtime as fully as deformable or dynamic convolutions; on scenes with highly nonstationary context, this can leave performance untapped. Third, while the U-shaped cascaded LUTs aggregate information across multi levels without resolution changes, they do not provide true multi-scale resizing, which can limit performance on scenes/objects that require explicit scale normalization.

\section{Conclusion}
\label{sec:conclusion}
We presented RFE-LUT, a look-up–table framework that reconciles large receptive fields with strict memory budgets by coupling lattice vector quantization (LVQ) with receptive-field expansion via irregular dilated convolutions and a U-shaped cascaded LUT design. The LVQ formulation maximizes table utility under fixed capacity, while IDC and U-LUTs enrich long-range context without incurring heavy compute. Across image restoration and segmentation benchmarks, RFE-LUT consistently surpasses prior LUT-based methods and narrows the gap to CNN baselines, all with substantially lower storage and latency—qualities that make it practical for real-time, resource-constrained deployment.
Looking ahead, we see opportunities to further compress tables (e.g., with advanced structured priors), automate dilation pattern search, and extend LVQ-aware training to broader architectures and tasks. We believe RFE-LUT provides a scalable foundation for fast visual inference where efficiency and accuracy must co-exist.

\bibliographystyle{IEEEtran}
\bibliography{rfe-lut}

\begin{IEEEbiography}
  [{\includegraphics[width=1in,clip,keepaspectratio]{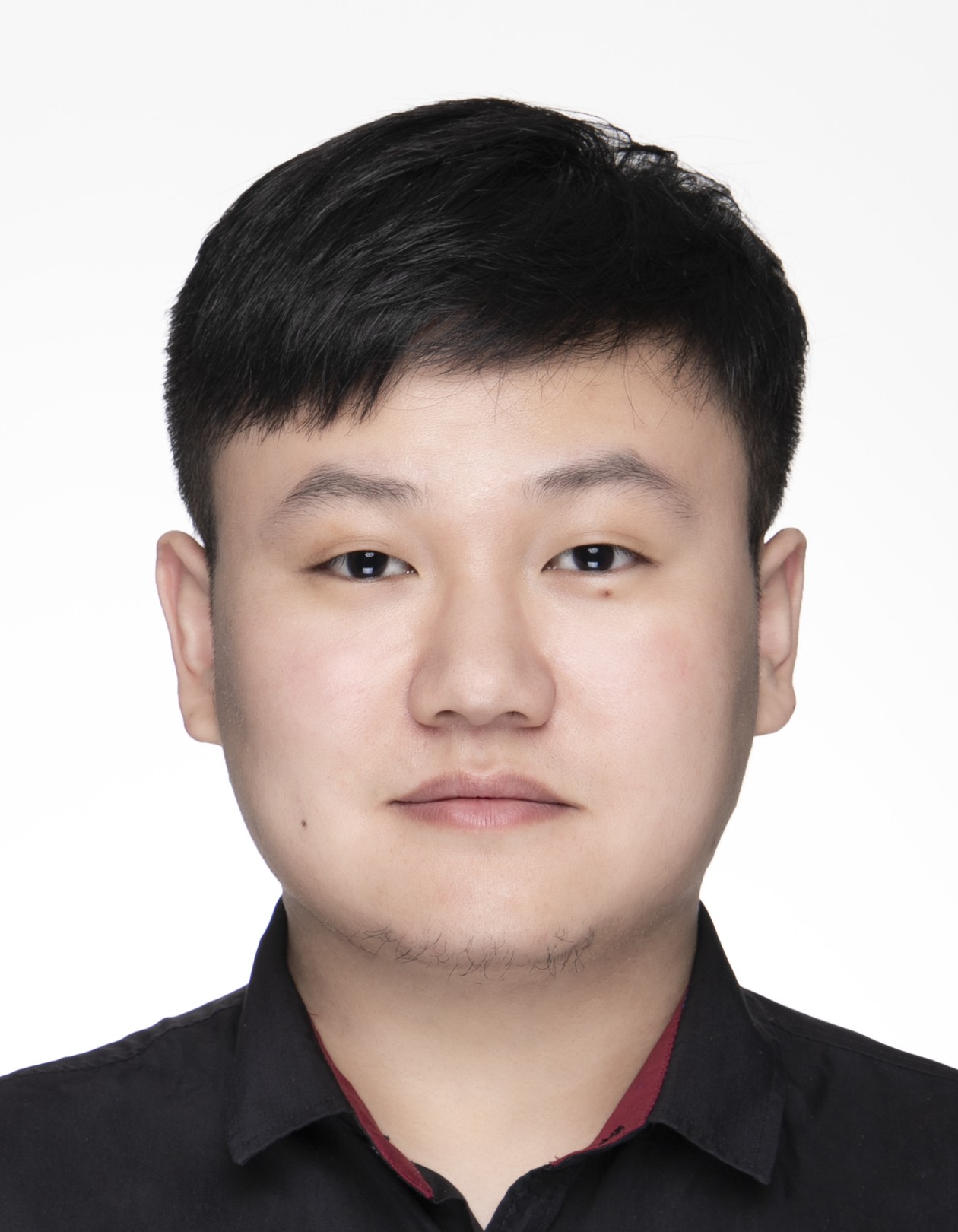}}]{Xi Zhang}
  (Member, IEEE) 
  received the B.Sc. degree in mathematics and physics basic science 
  from the University of Electronic Science and Technology of China, in 2015, and the Ph.D. degree 
  in electronic engineering from Shanghai Jiao Tong University, China, in 2022.
  He was a postdoctoral fellow at McMaster University (Mac), Canada, from July 2022 to August 2024. 
  He is currently a Research Scientist with the Alibaba-NTU Global e-Sustainability CorpLab (ANGEL) 
  at Nanyang Technological University (NTU).
  His current research focuses on Green AI, particularly on efficient model design, sustainable system architectures, and resource-aware compression techniques.  
  He is also interested in broader challenges in deep learning, including domain generalization and visual reasoning.
  He has published more than 10 papers at the TOP Journals and conferences 
  such as T-PAMI, T-IP, NeurIPS, CVPR, etc. 
\end{IEEEbiography}

\begin{IEEEbiography}
  [{\includegraphics[width=1in,clip,keepaspectratio]{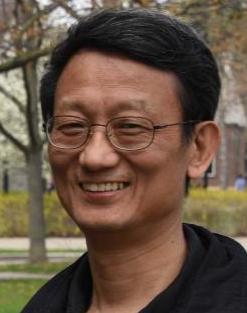}}]{Xiaolin Wu}
  (Life Fellow, IEEE) 
  received the B.Sc. degree in computer science 
  from Wuhan University, China, in 1982, and the Ph.D. degree in computer science 
  from the University of Calgary, Canada, in 1988. He started his academic career 
  in 1988. He was a Faculty Member with Western University, Canada, and New York 
  Polytechnic University (NYU-Poly). He is currently with McMaster University and 
  Southwest Jiaotong University. His research interests include image processing, 
  data compression, digital multimedia, low-level vision, and network-aware visual 
  communication. He has authored or co-authored more than 300 research articles and
  holds four patents in these fields. He served on technical committees for many 
  IEEE international conferences/workshops on image processing, multimedia, data 
  compression, and information theory. He was a past Associate Editor of IEEE 
  Transactions on Multimedia and IEEE Transactions on Image Processing.
\end{IEEEbiography}

\end{document}